\documentclass[conference]{IEEEtran}
\IEEEoverridecommandlockouts
\usepackage{cite}
\usepackage{amsmath,amssymb,amsfonts}
\usepackage{algorithm}
\usepackage{algorithmic}
\usepackage{multirow}
\usepackage{graphicx}
\usepackage{cite}
\usepackage{hyperref}
\usepackage{cleveref}
\usepackage{threeparttable}
\usepackage{booktabs}
\usepackage{textcomp}
\usepackage{xcolor}
\def\BibTeX{{\rm B\kern-.05em{\sc i\kern-.025em b}\kern-.08em
    T\kern-.1667em\lower.7ex\hbox{E}\kern-.125emX}}
\begin{document}

\title{
    SRAM-Based Compute-in-Memory Accelerator for Linear-decay Spiking Neural Networks
    \thanks{
        
    }
}

\author{
    \IEEEauthorblockN{
        Hongyang Shang$^{\dagger}$, 
        Shuai Dong$^{\dagger}$, 
        Yahan Yang,
        Junyi Yang,
        Peng Zhou$^{\ast}$,
        Arindam Basu$^{\ast}$
    }
    \IEEEauthorblockA{
        Department of Electrical Engineering, City University of Hong Kong, Hong Kong, China\\
    }
    \thanks{${\dagger}$Hongyang Shang and Shuai Dong contribute equally. $^{\ast}$Corresponding author (e-mail: pengzhou@cityu.edu.hk, arinbasu@cityu.edu.hk). This work was supported by the Research Grants Council of the HK SAR, China (Project No. CityU 11212823 and CityU 11208125).
}
}

\maketitle
\begin{abstract}
Spiking Neural Networks (SNNs) have emerged as a biologically inspired alternative to conventional deep networks, offering event-driven and energy-efficient computation. However, their throughput remains constrained by the serial update of neuron membrane states. While many hardware accelerators and \textit{Compute-in-Memory (CIM)} architectures efficiently parallelize the synaptic operation ($\mathbf{W}\times\mathbf{I}$) achieving $\mathcal{O}(1)$ complexity for matrix–vector multiplication, the subsequent state update step still requires $\mathcal{O}(N)$ time to refresh all neuron membrane potentials. This mismatch makes state update the dominant latency and energy bottleneck in SNN inference. 
To address this challenge, we propose an \textit{SRAM-based CIM} for SNN with \textit{Linear Decay Leaky Integrate-and-Fire (LD-LIF) Neuron} that co-optimizes algorithm and hardware. At the algorithmic level, we replace the conventional exponential membrane decay with a linear decay approximation, converting costly multiplications into simple additions while accuracy drops only around 1\%. At the architectural level, we introduce an in-memory parallel update scheme that performs in-place decay directly within the SRAM array, eliminating the need for global sequential updates.
Evaluated on benchmark SNN workloads, the proposed method achieves a $1.1 \times$ to $16.7 \times$ reduction of SOP energy consumption, while providing $15.9 \times$ to $69 \times$ more energy efficiency, with negligible accuracy loss relative to original decay models. This work highlights that beyond accelerating the $\mathbf{W}\times\mathbf{I}$ computation, optimizing state-update dynamics within CIM architectures is essential for scalable, low-power, and real-time neuromorphic processing.

\end{abstract}


\section{Introduction}
Spiking Neural Networks (SNNs) have emerged as a promising paradigm, achieving high energy efficiency and temporal sparsity. Neuromorphic hardware accelerators achieve ultra-low power such as Loihi/Loihi2 ~\cite{davies2018loihi, intelloihi2}, TrueNorth/NorthPole~\cite{akopyan2015truenorth,modha2023neural}, Spinnaker1/2~\cite{painkras2013spinnaker, mayr2019spinnaker}. To further minimize the von Neumann bottleneck, Compute-in-Memory (CIM) architectures have demonstrated that synaptic operations can be executed directly within memory arrays, providing massive parallelism and near $\mathcal{O}(1)$ latency per timestep ~\cite{yu2021compute,yang2025efficient}.

However, each neuron in an SNN must also update its membrane potential through a decay-and-accumulation process, which remains a key bottleneck. Unlike highly parallel synaptic accumulation in CIM arrays, they require sequential neuron-state access, leading to $\mathcal{O}(N)$ latency and high memory-access energy. This mismatch between highly parallel synaptic computation and serial state update has become a dominant limitation in large-scale SNN inference.

To address this challenge, we propose an \textit{SRAM-based CIM linear-decay architecture} that co-optimizes algorithm and hardware. At the algorithmic level, the exponential membrane decay used in conventional neuron models is replaced by a simplified linear decay approximation, eliminating the need for multipliers while maintaining comparable accuracy. At the architectural level, the proposed CIM macro integrates decay and update operations directly into the memory array, enabling in-place and parallel updates of membrane states. Among various memory technologies explored for CIM, including RRAM, PCM, MRAM, FeFET, etc, \textit{SRAM} offers the best balance between performance, reliability, and CMOS compatibility. Its high read–write speed, robust stability, and proven scalability make it ideal for fine-grained and frequent state updates required in SNNs ~\cite{yu2021compute}.

\begin{figure}[htbp]
\centerline{\includegraphics[width=0.48\textwidth]{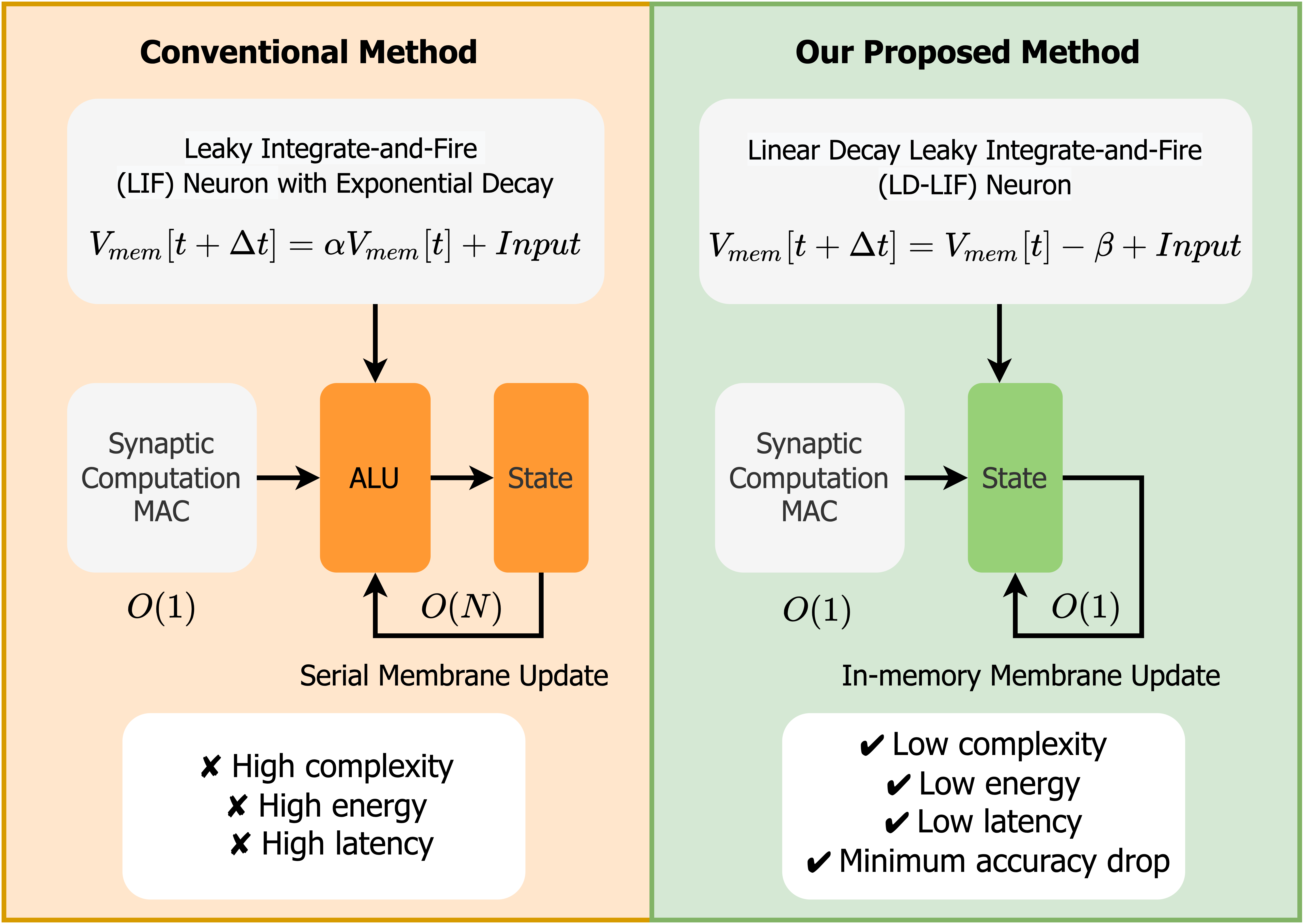}}
\caption{Comparison between the conventional LIF model with exponential decay and the proposed LD-LIF implemented in SRAM-based CIM.}
\label{overview}
\end{figure}

Fig.~\ref{overview} illustrates the conceptual comparison between the conventional leaky integrate-and-fire (LIF) neuron with exponential decay and our proposed linear-decay LIF (LD-LIF) implemented in an SRAM-based CIM architecture. Experimental results show that the proposed architecture significantly reduces latency and energy consumption with negligible accuracy degradation, offering a scalable pathway toward high-efficiency neuromorphic CIM processors.

\section{Methodology}
\subsection{Linear Decay Leaky Integrate-and-Fire (LD-LIF) Neuron}

The membrane dynamics of a LIF neuron can be described as
\begin{equation}
\tau_m \frac{dV_{\text{mem}}}{dt} = -V_{\text{mem}} + \sum W \cdot S_{\text{in}},
\label{lif_ct}
\end{equation}
where $S_{\text{in}}$ denotes the pre-synaptic spike input and $W$ is the synaptic weight.

If the neuron starts from an initial potential $V_0$ and receives no further input ($S_{\text{in}}=0$), the solution of \eqref{lif_ct} is
\begin{equation}
V_{\text{mem}}(t) = V_0 e^{-t / \tau_m}
\end{equation}

Applying the Forward Euler method to \eqref{lif_ct} yields a discrete-time update:
\begin{equation}
V_{\text{mem}}[t{+}\Delta t] 
= \left(1 - \frac{\Delta t}{\tau_m}\right) V_{\text{mem}}[t]
+ \frac{\Delta t}{\tau_m} \sum W \cdot S_{\text{in}}[t].
\end{equation}

For small $\Delta t$, this is approximately equivalent to the exponential form with $\alpha \approx e^{-\Delta t / \tau_m}$. 
However, realizing this exponential operation in hardware requires multipliers or lookup tables, increasing both power and area cost.

\begin{figure}[t]
\centerline{\includegraphics[width=0.45\textwidth]{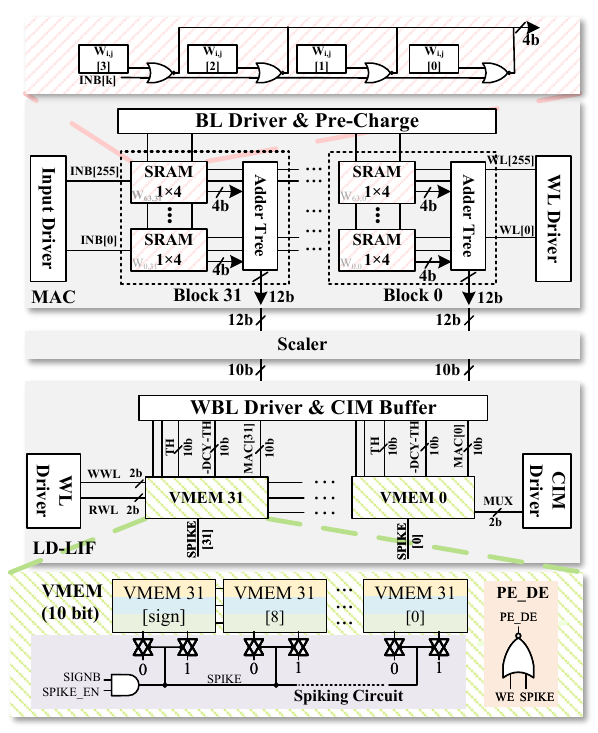}}
\caption{Hardware Architecture of SRAM-based In-Memory Computing with Linear Decay.}
\label{total_arch}
\end{figure}

\subsection{Unit Cell for Membrane Potentials}
\begin{figure}[t]
\centerline{\includegraphics[width=0.48\textwidth]{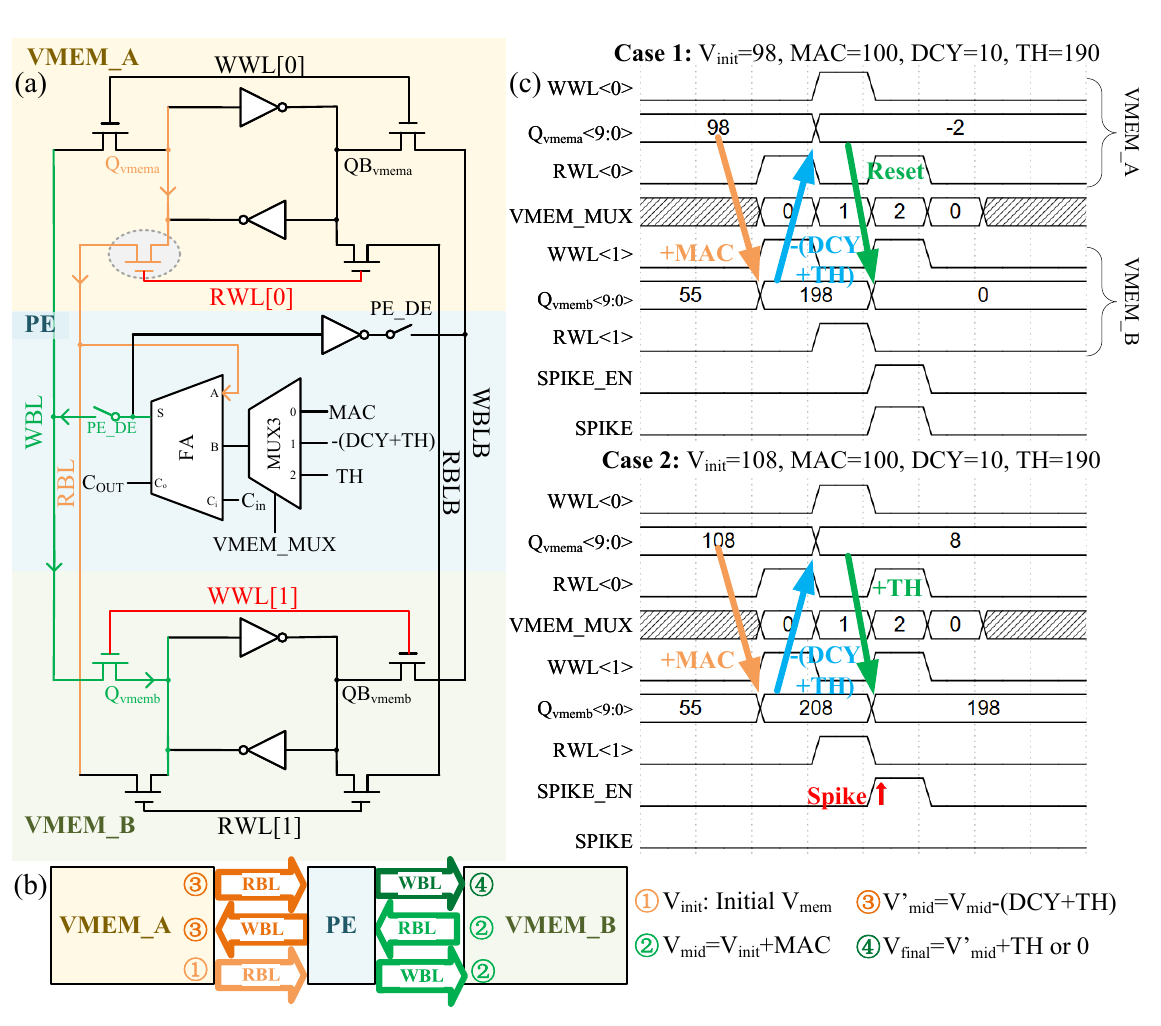}}
\caption{Schematic and operation of the membrane voltage cell. (a) Circuit of the membrane voltage cell. (b) Data flow for the membrane voltage update process. (c)Waveform of membrane voltage cell operation.}
\label{vmem_cell}
\end{figure}

To simplify implementation, we adopt a \textit{linear decay approximation}:
\begin{equation}
V_{mem}[t+1] = V_{mem}[t] - \beta + W \cdot \text{Input}[t],
\end{equation}
where $\beta$ is a fixed decrement representing membrane leakage per timestep. This substitution converts the exponential leakage into an simple add/subtract process, removing multipliers and enabling direct in-memory implementation. Prior studies have shown that simplified decay functions, such as linear or piecewise-linear forms, preserve SNN accuracy and robustness~\cite{eissa2021hardware,liu2020fpt,chowdhury2021towards}. 

While FPT-Spike~\cite{liu2020fpt} adopted a linearly decayed spike kernel to simplify the exponential post-synaptic response, our work extends this concept to the membrane update dynamics, replacing the exponential leakage term with a constant linear decrement. Both approaches share the same motivation of linearizing the exponential decay for hardware efficiency, but operate at different levels—kernel shaping versus membrane potential update. The proposed linear decay model thus provides a hardware-friendly yet dynamically effective foundation for efficient CIM SNN implementation.

\subsection{SNN Learning Algorithm with LD-LIF Neuron}

To preserve accuracy, the linear decay method incorporates a learnable decay parameter \(\beta\), which is essential for adjusting the dynamics of the membrane potential for different layers. Each LIF layer shares a single \(\beta\) across all output neurons, to simplify the hardware implementation. During the forward pass, the membrane potential from the previous time step, \(V(t-1)\), is first reduced by the learnable decay parameter \(\beta\), and then updated by adding the input \(I(t)\) from the Multiply-Accumulate (MAC) output, resulting in the updated \(V(t)\). If \(V(t)\) exceeds the threshold \(\theta\), the spike \(S(t)\) is triggered, and \(V(t)\) is reset to zero to prepare for the next time step. In the backward pass, the \(\beta\) is updated using gradient descent. By learning an optimal \(\beta\) for each layer, the network can better capture the temporal dependencies in the data and enhance its ability to make accurate predictions, leading to better overall performance in a variety of applications.


\subsection{Hardware Architecture for SNN}

The architecture of the linear decay SNN accelerator is shown in the Fig. \ref{total_arch}. This architecture mainly consists of three parts: the MAC module, the Scaler module, and the LD-LIF module. The MAC contains 32 SRAM blocks, and each SRAM block consists of 256 rows × 4 SRAM units, where 4 SRAM units store 4-bit weights, along with an adder tree. As shown in the pink background subgraph, each 1x4 SRAM is made up of 4 regular 6T SRAM units and 4 NOR gates. One input of the NAND gate is an inverted bit of weight (WB), and the other input is INB, providing the 4-bit multiplication result. Since the input to the SNN is either 1 or 0, the input is a single-bit value. The MAC result is sent to a scaler module, which aligns it with the LIF module’s 10-bit VMEM, ensuring consistent scaling between the MAC output and $V_{mem}$. The LIF module includes 32 VMEM submodules, with each VMEM representing a 10-bit membrane potential. Additionally, each VMEM submodule includes processing an element disenabling circuit (PE\textunderscore DE with orange background) and a spiking circuit (purple background).

\begin{figure}[t]
\centerline{\includegraphics[width=0.5\textwidth]{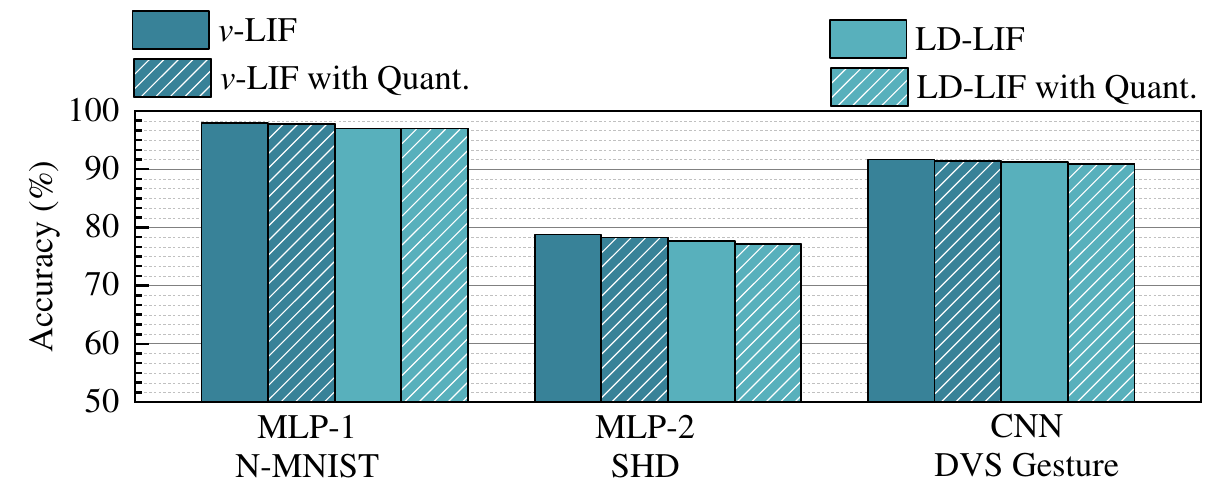}}
\caption{Performance comparison of Baseline, Linear Decay, and Quantization methods among SHD, NMNIST and DVS Gesture datasets.}
\label{fig:acc}
\end{figure}

\begin{figure}[t]
\centerline{\includegraphics[width=0.5\textwidth]{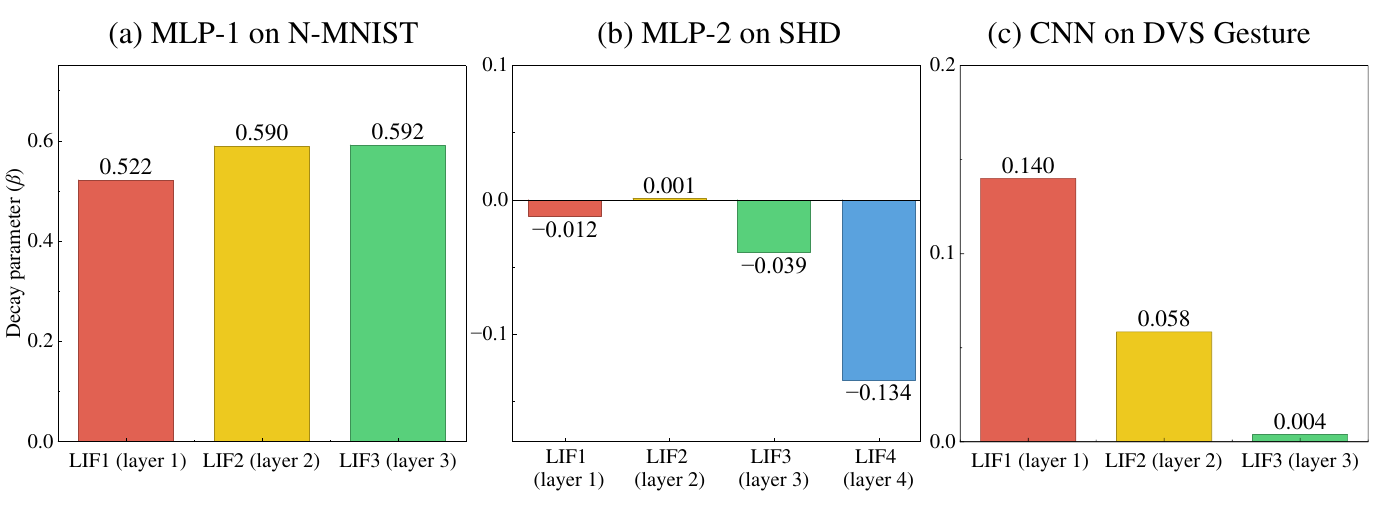}}
\caption{The learned linear decay $\beta$ parameters among three networks: (a) MLP-1 on N-MNIST, (b) MLP-2 on SHD, and (c) CNN on DVS Gesture.}
\label{fig:beta}
\end{figure}

\begin{figure}[t]
\centerline{\includegraphics[width=0.5\textwidth]{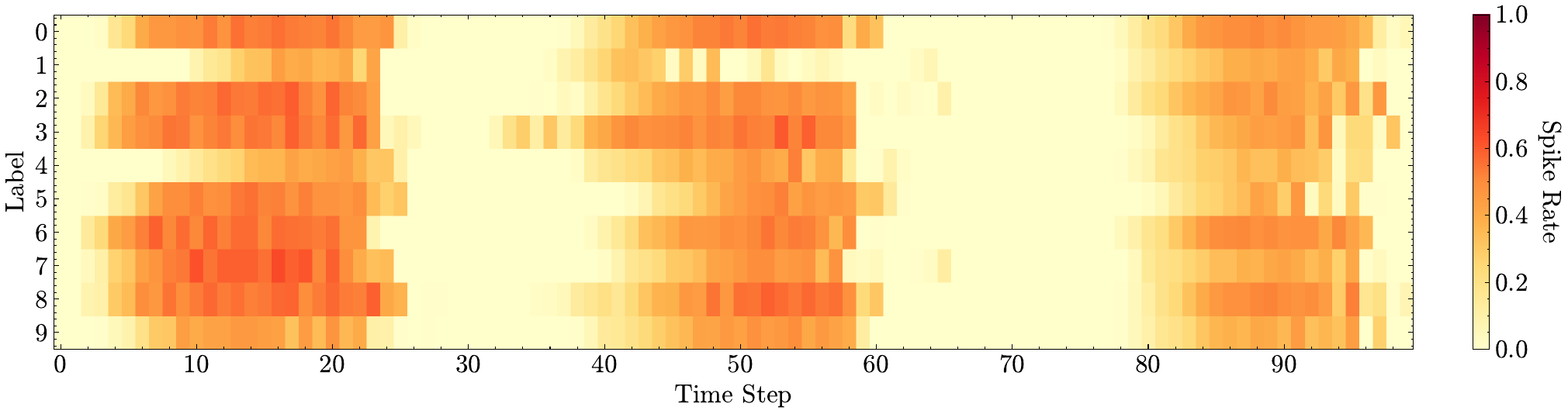}}
\caption{Spiking rate in the second LD-LIF layer of MLP-1 on the N-MNIST dataset with labels 0-9 (representing 10 N-MNIST classes).}
\label{fig:spiking_rate}
\end{figure}

As shown in Fig. \ref{vmem_cell}(a), a single VMEM cell consists of two 8T SRAM blocks (VMEM\textunderscore A and VMEM\textunderscore B) and a processing element (PE). The PE comprises a full adder (FA) and a 3-to-1 multiplexer (MUX3), one inverter and two switches. The 8T SRAM provides separate write (WWL/WBL) and read (RWL/RBL) paths. Note that the RBL here is not used for conventional differential SRAM readout. Therefore, neither precharge circuits nor sense amplifiers are required. Instead, an NMOS pass transistor directly forwards the stored voltage in VMEM\textunderscore A or VMEM\textunderscore B to the input of the FA. To mitigate threshold-voltage drop, the NMOS devices highlighted with the gray dashed circle are implemented using the lvt type.

The three inputs of the MUX3, selected by $VMEM\textunderscore MUX$, are the multiply-accumulate results ($MAC$), the negative sum of decay and threshold ($-(DCY+TH)$), and the threshold $TH$. The MUX3 output drives the $B$ port of the FA. The $A$ port of the FA is supplied by the $RBL$: when $RWL[0]$ is asserted high, $Q_{vmema}$ (from VMEM\textunderscore A) is read; when $RWL[1]$ is asserted high, $Q_{vmemb}$ (from VMEM\textunderscore B) is read. The carry-in ($C_{in}$) and carry-out ($C_{out}$) are connected in the usual ripple manner to the adjacent lower and higher bits, respectively. The two switches are used to disconnect the output port $S$ of FA to the $WBL$ and $WBLB$ in the writing mode and spiking mode. 

In Fig. \ref{vmem_cell}(b), the stored membrane potential $V_{mem}$ flows between VMEM\textunderscore A and VMEM\textunderscore B in a ping–pong manner, completing the update of $V_{mem}$ within three clock cycles. The initial value $V_{init}$ resides in VMEM\textunderscore A. In the first cycle, $V_{init}$ is processed by the PE to add the MAC value, yielding the intermediate value $V_{mid}$, which is written to VMEM\textunderscore B via $WBL$ within the same cycle. In the second cycle, $V_{mid}$ is read from  VMEM\textunderscore B and passed through the PE to subtract the sum of DCY and TH, producing $V'_{mid}$, which is then stored back in VMEM\textunderscore A. The sign bit of $V'_{mid}$ indicates the state of spike. If the sign bit is 1, then $V'_{mid}$ is negative, meaning $V_{init} + MAC - DCY$ is smaller than $TH$. Otherwise, $V_{init} + MAC - DCY$ is bigger than $TH$ and there should be a spike. In the third cycle, $WBL$ is either driven by FA or spiking circuit and update $V_{final}$ in VMEM\textunderscore B. If it is the case of no spiking, $V'_{mid}$ is processed once more by the PE with MUX3 selecting TH and $V_{final}$ = $V'_{mid}$ + TH. In the other case of there is a spiking, the final value written back is $V_{final}$ = 0. As illustrated, we changed the sequence of membrane potential updates. By subtracting the decay value together with the threshold and leveraging the sign bit for comparison, this approach eliminates the overhead of a comparison circuit. The detailed waveform of the signals are shown in Fig.\ref{vmem_cell}(c). 




\section{Results}
\subsection{Algorithm Results}
We evaluated the LD-LIF method on three networks: MLP-1 (2312-256-128-10) on N-MNIST, MLP-2 (140-128-128-128-10) on Spiking Heidelberg Digits (SHD), and CNN (conv1-conv2-linear) on DVS Gesture. The vanilla LIF (v-LIF) decay method is used as the baseline, achieving accuracies of 97.95\%, 78.80\%, and 91.67\%, respectively.

Fig. \ref{fig:acc} compares the performance of the proposed LD-LIF with the v-LIF. LD-LIF results in only a slight reduction in accuracy compared to v-LIF, with decreases of 0.96\%, 1.11\%, and 0.38\%, demonstrating the effectiveness of our method. Additionally, we investigated the impact of weight quantization. For MLP-1, 3-bit quantization is applied, whereas MLP-2 and the CNN employ 4-bit quantization. The quantization losses of v-LIF and LD-LIF remain comparable, ranging from 0.02\% to 0.37\% for LD-LIF and 0.13\% to 0.41\% for v-LIF.

To further investigate the effectiveness of the learned linear decay, the decay parameters $\beta$ are presented in Fig. \ref{fig:beta}. For the MLP-1 network on N-MNIST, the decay parameters are higher, ranging from 0.522 to 0.592, compared to the MLP-2 on SHD (ranging from -0.134 to 0.001) and the CNN on DVS Gesture (ranging from 0.004 to 0.140). Notably, the linear decay in MLP-2 exhibits negative values, indicating that the membrane potential increases with each time step ($Vmem (t)-\beta$). This suggests that LIF layers with negative $\beta$ tend to spike more frequently, highlighting the distinct impact of linear decay on spiking behavior.

To investigate the spiking behavior of the linear decay mechanism, we randomly selected 10 samples from different classes of the N-MNIST dataset for the MLP-1 network. As shown in Fig. \ref{fig:spiking_rate}, the spiking rates across 100 time steps in the second LD-LIF layer is presented. The spiking rate reflects the number of neurons that spike within each layer at a given time step. The results show that the LD-LIF layer tends to spike around three key time points: $t$=15, $t$=50, and $t$=90, indicating an amplification of spiking activity as the signal progresses through the layers.

\begin{figure}[t]
\centerline{\includegraphics[width=0.5\textwidth]{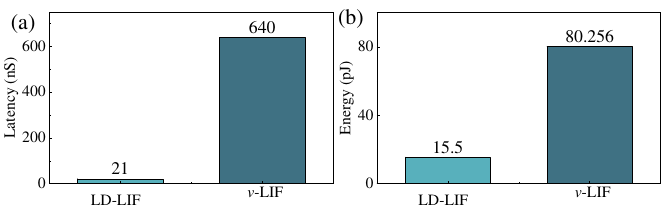}}
\caption{Comparison of latency and energy consumption between the LD-LIF and v-LIF models).}
\label{fig:lif_comp}
\end{figure}


\begin{table}[]

\begin{center}
\begin{threeparttable}
\caption{Comparison with state-of-the-arts}

\begin{tabular}{|c|c|c|c|c|}
\hline
&\begin{tabular}[c]{@{}c@{}}JSSC’24\\ \cite{zhang2024anp}\end{tabular}  
&\begin{tabular}[c]{@{}c@{}}VLSI’23\\ \cite{huo2023anp}\end{tabular} 
&\begin{tabular}[c]{@{}c@{}}TCASAI’25\\ \cite{kim2025energy}\end{tabular}
& Prop.\\ \hline
Tech. (nm)                          & 28        & 28         & 180               & 65   \\ \hline
Supp. (V)                           & 0.56-0.9  & 0.55-0.9  & 1.8                & 0.9$^{1}$   \\ \hline
Freq. (MHz)                         & 40-210    & 30-240     & 150               & 142\\ \hline
Computing                           & Digital   & Digital    &Mixed-Signal       &Digital   \\ \hline
W. bit \#                           & 8-10      & 4          & 3                 &4    \\ \hline
Neuron type                         & LIF       & LIF        &Stochastic LIF     &LD-LIF  \\ \hline
SOP energy (pJ/SOP)                 & 1.5       & 1.04       &0.1                &0.09   \\ \hline
TOPS/W                              & 0.3       & 0.96       &1.3                &20.7   \\ \hline
\end{tabular}

\begin{tablenotes}
      \item $^1$ The operating voltage for the LD-LIF module is 0.9 V, with the MAC and Scaler modules operating at 0.72 V and 1.2 V, respectively.
    \end{tablenotes}

  \end{threeparttable}
\end{center}
\label{tab1}
\end{table}

\subsection{Latency and Energy Comparisons for LD-LIF Module}
The hardware results are simulated under TSMC 65 nm process. Fig. \ref{fig:lif_comp} compares the latency and energy efficiency of our proposed LD-LIF module against a digital implemented v-LIF. Regarding latency (Fig. \ref{fig:lif_comp}(a)), our architecture completes the update of all 32 membrane potentials in parallel within just 3 clock cycles. At a 0.9 V supply voltage, the slowest single update takes 7 ns, resulting in a total latency of only 21 ns. In contrast, a conventional serial architecture operating at 200 MHz requires at least 4 cycles per membrane potential, leading to a total latency approximately 30 times higher than our design. In terms of energy consumption (Fig. \ref{fig:lif_comp}(b)), our design achieves an approximately 5.2 times improvement in energy efficiency comparing with conventional digital LIF, whose power is from DC synthesis. These results demonstrate the significant advantages of our approach in both latency and power consumption.

\subsection{Comparison with Other Existing Works}

To further highlight the advantages of our approach, we compared it with three existing SNN hardware implementations.The power consumption of the MAC module is referenced from \cite{chih202116}, and scaled to the 65 nm technology node using the formula provided in \cite{zhou2025layer}. As shown in Table I, the proposed method achieves 0.09 pJ/SOP energy per synaptic operation (SOP) and 21.6 TOPS/W energy efficiency. Compared to other works, this results in a reduction of SOP energy consumption by a factor of $1.1 \times$ to $16.7 \times$, while providing $15.9 \times$ to $69 \times$ greater energy efficiency.

\section{Conclusion}

In conclusion, this work addresses the throughput and energy bottlenecks in SNN by proposing an SRAM-based CIM linear decay architecture. By replacing the conventional exponential membrane decay with a linear decay approximation, we significantly reduce the computational complexity from expensive multiplications to efficient additions, while preserving accuracy. On the hardware side, we introduce an in-memory parallel update scheme that performs in-place decay directly within the SRAM array, thereby eliminating the need for global sequential updates.
The proposed method was evaluated on benchmark SNN workloads and demonstrated significant improvements, achieving a reduction in SOP energy consumption by a factor of $1.1 \times$ to $16.7 \times$, and a substantial increase in energy efficiency by a factor of $15.9 \times$ to $69 \times$, with negligible accuracy loss compared to traditional decay models. These results underscore the importance of optimizing state-update dynamics within CIM architectures, in addition to accelerating the synaptic operation. This approach provides a scalable, low-power, and real-time solution for neuromorphic processing.

\clearpage
\clearpage

\bibliography{refs.bib}
\bibliographystyle{IEEEtran} 

\end{document}